\documentclass[10pt,twocolumn,letterpaper]{article}

\usepackage{cvpr}
\usepackage{times}
\usepackage{epsfig}
\usepackage{graphicx}
\usepackage{amsmath}
\usepackage{amssymb}
\usepackage{enumitem}

\usepackage[breaklinks=true,bookmarks=false]{hyperref}

\cvprfinalcopy 

\def\cvprPaperID{1716} 

\ifcvprfinal\fi
\begin{document}

\title{Longitudinal Face Modeling via \\
	Temporal Deep Restricted Boltzmann Machines}

\author{Chi Nhan Duong $^{1}$, Khoa Luu $^{2}$, Kha Gia Quach $^{1}$ and Tien D. Bui $^{1}$\\
	$^{1}$ Computer Science and Software Engineering, Concordia University, Montr\'eal, Qu\'ebec, Canada\\
	$^{2}$ CyLab Biometrics Center and the Department of Electrical and Computer Engineering, \\ Carnegie Mellon University, Pittsburgh, PA, USA\\
	{\tt\small $^{1}$\{c\_duon, k\_q, bui\}@encs.concordia.ca, $^{2}$kluu@andrew.cmu.edu}
}

\maketitle

\begin{abstract}
  Modeling the face aging process is a challenging task due to large and non-linear variations present in different stages of face development.
  This paper presents a deep model approach for face age progression that can efficiently capture the non-linear aging process and automatically synthesize a series of age-progressed faces in various age ranges. In this approach, we first decompose the long-term age progress into a sequence of short-term changes and model it as a face sequence. The Temporal Deep Restricted Boltzmann Machines based age progression model together with the prototype faces are then constructed to learn the aging transformation between faces in the sequence. In addition, to enhance the wrinkles of faces in the later age ranges, the wrinkle models are further constructed using Restricted Boltzmann Machines to capture their variations in different facial regions. The geometry constraints are also taken into account in the last step for more consistent age-progressed results.
  The proposed approach is evaluated using various face aging databases, i.e. FG-NET, Cross-Age Celebrity Dataset (CACD) and MORPH, and  our collected large-scale aging database named AginG Faces in the Wild (AGFW).
  In addition, when ground-truth age is not available for input image, our proposed system is able to automatically estimate the age of the input face before aging process is employed.
\end{abstract}

\begin{figure}[t]
	\begin{center}
		\includegraphics[width=8.0cm]{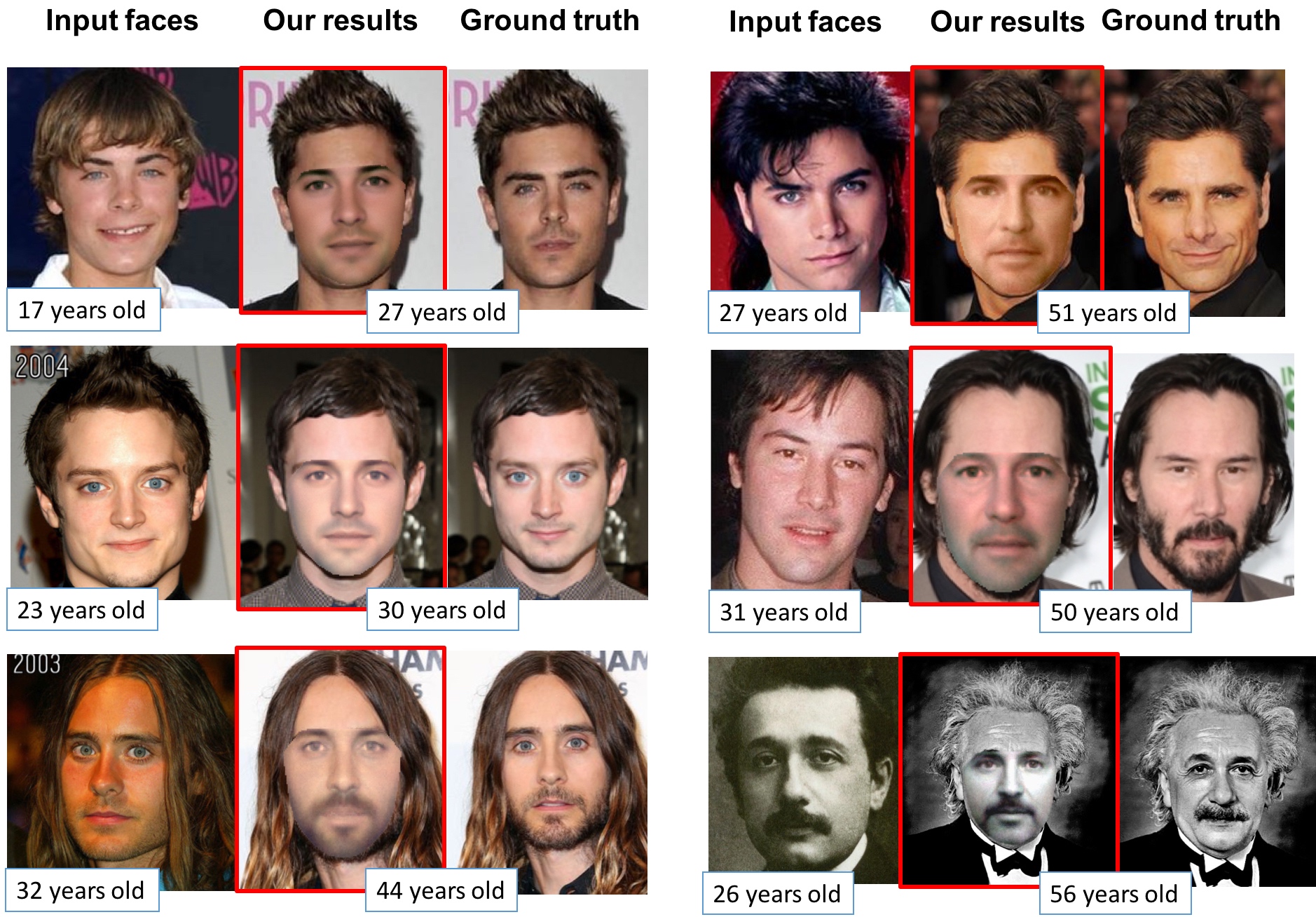}
	\end{center}
	\caption{Examples of age progression using our proposed approach. Each subject has three images: the input image (left), the synthesized age-progressed face (middle), and the ground truth (right). Our system also can predict the ages of input faces in case these ground-truths are not available.
	}
	\label{fig:AgeProgressionCelebrities}
\end{figure}
\begin{figure*}[!t]
	\begin{center}
		\includegraphics[width=15cm]{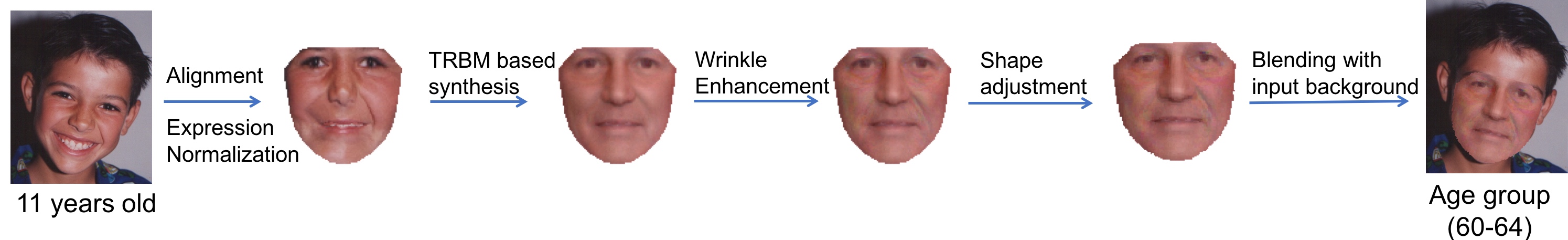}
	\end{center}
	\caption{Processing steps of our proposed method to synthesize the face at ages of 60s given a face at age of 10-14}
	\label{fig:ProcessingStep}
\end{figure*}
\setlength{\textfloatsep}{0.7cm}

\section{Introduction}

Face age progression presents the capability to predict future faces of an individual in input photos. In most cases, there is only one photo of that individual and we have to predict the future faces, i.e. age progression, or construct the former faces, i.e. age regression or deaging, of that subject \cite{Albert2007a}.
Face aging can find its origins from missing children when police require age progressed pictures. This problem is also applicable in cases of wanted fugitives where face age progression is also required. The predominant approach to aging pictures
involves the use of forensic artists \cite{Taylor2000a}. Although forensic artists are trained in the anatomy and geometry of faces,
they still can suffer from psycho-cognitive bias that may affect their interpretation of the source face data. In addition, an age-progressed image can differ significantly from one forensic artist to the next. Manual age progression usually takes lots of time and requires the work of numerous professional forensic artists. Therefore, automatic and
computerized age-progression systems are important. Their applications range from very sensitive national security problems to tobacco or alcohol stores/bars to control the patron's age and cosmetic studies against aging.

Synthesizing plausible faces of individuals at different stages in their life is an extremely challenging task, even for human, due to several reasons. Firstly, human face aging is a complicated process since people usually age in different ways. It is non-deterministic and greatly depends on intrinsic factors, i.e. gender, ethnicity and heredity. Moreover, extrinsic factors, i.e. environment, living styles and smoking, have also created various effects to the facial changes and resulted in large aging variations even between people in the same age group. Secondly, facial shapes and textures dramatically change over the long periods.
Thirdly, it is very hard to collect a longitudinal face age database that is generative enough to learn an aging model. Currently existing aging databases in the research community are small or unbalanced among genders, ethnicities and age groups. In addition, they are usually mixed with other variations, e.g. expressions and illuminations.

Automatic face age progression has attracted huge interest from the computer vision community in recent years.
There are numerous efforts to model the longitudinal aging process presented in computer vision literature \cite{geng2007automatic, lanitis2002toward, patterson2006automatic, suo2012concatenational, kemelmacher2014illumination}. In most conventional methods, linear models, e.g. Active Appearance Models (AAMs) and 3D Morphable Model, are usually adopted to interpret the geometry and appearance of the faces before the aging rules are learned.
However, the face aging variations are not only large but also non-linear. It apparently violates the assumption of linear models. Therefore, these age-progression methods meet a lot of difficulties and limitations to interpret these non-linear aging variations.

Recently, Temporal Restricted Boltzmann Machines (TRBM) \cite{sutskever2007learning,taylor2006modeling,zeiler2011facial} have gained attention significantly as one of the probabilistic models to accurately model complex time-series structure while keeping the inference tractable.
As an extension of Restricted Boltzmann Machines (RBM), the structure of TRBM consists of further directed connections from previous states of visible and hidden units. By this way, the short history of their activations can act as ``memory'' and is able to contribute to the inference step of visible units.
In this structure, multiple factors are learned and interacted to efficiently explain the temporal data. Therefore, TRBM provides the ability to extract more complicated and nonlinear structures in time series data.

This work presents a novel deep model based approach to face age progression.
Instead of synthesizing faces directly from long periods, the long-term aging process is considered as a set of short-term changes and presented using a sequence of faces.
The TRBM based model is then constructed to capture the aging transformation between consecutive faces in the sequence.
In addition, to enforce the model on the capabilities of aging variations, a set of reference faces that are mainly different in age conditions is generated and incorporated into the model.
Then, a set of RBMs based wrinkle models is developed to enhance the wrinkle details in these aging faces.
Finally, the facial geometric information of each age group is extracted and adopted to adjust the face shapes.
Figure \ref{fig:ProcessingStep} illustrates the main processing steps of our proposed system.
\begin{figure*}[t]
	\begin{center}
		\includegraphics[width=17cm]{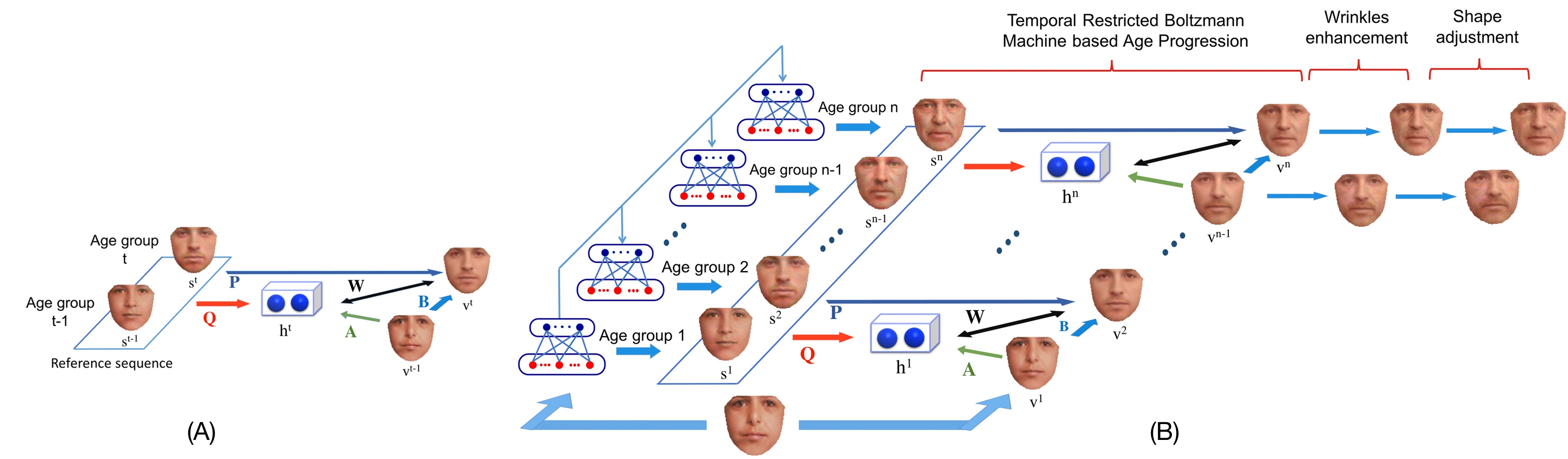}
	\end{center}
	\caption{The proposed age progression approach: (A) Temporal Restricted Boltzmann Machines for learning aging transformation in a single node; (B) The proposed system using multiple nodes; wrinkle enhancement and shape adjustment.}
	\label{fig:TRBM_whole_system}
\end{figure*}
The novelties of our approach are :

\begin{itemize}[leftmargin=*]
	\itemsep0em
	\item The face structure and specific aging features presented in each age group are modeled using RBM. Compared to other linear models, the use of RBMs can help to better interpret the non-linear variations and produce faces with more aging details. In addition, the high-level features extracted from hidden layer can be transferred between RBMs of different age groups for reconstructing a reference face sequence that can benefit the learning process.
	
	\item Together with the reference sequence,
	the proposed TRBM based model provides an efficient way to capture the aging transformation between faces in different age groups. Similar to RBM, TRBM is more advanced in interpreting the complex and non-linear aging process.

	\item Far apart from previous approaches where wrinkles are cloned from an average face or the closest faces of each age group, we propose a machine learning based approach to learn these aging rules, i.e. construct a set of RBMs based wrinkle models for every age group. In this way, the method is able to learn their distributions and generate synthetic wrinkles by sampling from these distributions.
	As a result, our model is more flexible in producing more wrinkle types.

	\item 
	The geometric differences between face shapes in every age group are also taken into account in our system.
	\item A large-scale dataset named AginG Faces in the Wild (AGFW) is collected for analysing the aging effects.
	
\end{itemize}
	\section{Related Work}
	Generally, previous age progression approaches can be divided into two groups, i.e. the \textit{anthropology approach} and \textit{example-based approach}.
	
	In the first group, the main idea is to simulate the biological structure and aging process of facial features such as muscles and facial skins based on theories from anthropometric studies \cite{bando2002simple, berg2003aging, boissieux2000simulation}.	
	Inspiring from the `revise'  cardioidal strain transformation, Ramanathan et al. \cite{ramanathan2006modeling} proposed a physiological craniofacial growth model for age progression.
	Ramanathan et al. \cite{ramanathan2008modeling} later introduced an aging model that incorporates both shape and texture variation models.
	To simulate the geometry changes, the shape transformation models are designed to capture the aging variations of three facial muscles.
	For the texture model, an image gradient based transformation function is adopted to characterize the facial wrinkles and skin artifacts. 
	
	In the second group, a straightforward idea is to use the age prototypes \cite{rowland1995manipulating} defined by the average faces of people in the same age group. Then the age-progressed faces can be produced by adding the differences between the prototypes of the target and the query age groups to the input face.
	In recent work of Kemelmacher-Shlizerman et al. \cite{kemelmacher2014illumination}, the authors extended this idea with a large-scale collection of images  for age prototypes construction. Then illumination normalization and subspace alignment technique are proposed to better handle images with various lighting conditions.
	Another direction
	is to represent a face as a set of parameters and learning aging functions from the relationships between these facial parameters and age label.
	Lanitis et al. \cite{lanitis2002toward} proposed to use AAMs parameters and introduced several aging functions to model both generic and specific aging processes.
	Pattersons et al. \cite{patterson2006automatic} also used AAMs and aging function in their system. However, they put more efforts on simulating the adult aging stage. The genetic facial features of siblings and parents were also incorporated to age progression in \cite{luu2009Automatic}.
	
	Geng et al. \cite{geng2007automatic} proposed an AGing pattErn Subspace (AGES) approach for both age estimation and age synthesis. 
	Tsai et al. \cite{tsai2014human} then extended the AGES with the guidance faces corresponding to the subject's characteristics for more stable results.
	Suo et al. \cite{suo2010compositional} proposed to decompose a face into smaller components (i.e. eyes, mouth, etc.) and learning the aging process for each component.
	A three-layer And-Or graph is adopted for face representation. Then the changes in face aging are modeled by a Markov chain on parse graphs.
	Similarly, in \cite{suo2012concatenational}, Suo et al. further employed this decomposition strategy in temporal aspects where long-term evolution of the graphical representation is learned by connecting sequences of  short-term patterns.

	\section{Our Proposed Age Progression Approach}
	
	Our proposed age progression system (as shown in Figure \ref{fig:TRBM_whole_system}(B)) consists of five main steps: (1) Preprocessing, (2) Reference sequence generation; (3) Texture age progression; (4) Wrinkles enhancement; and (5) Shape adjustment.

	\subsection{Data Collection} \label{sec:DataCollection}
	In order to train the model and	to analyze the aging effects, a large-scale dataset named AginG Faces in the Wild (AGFW)\footnote{This dataset will be published online for later research uses.} is first collected.	Moreover, to ensure the consistency of the collected data, the tag names and the age-related information of these images are also considered. The resulting dataset consists of 18,685 images with the age ranging from 10 to 64 years. It is then decomposed into 11 age groups with the age span of 5 years.
	On average, each age group consists of 1700 images of different people in the same age group.
	The Productive Aging Laboratory (PAL) Face database \cite{minear2004lifespan} is also included in our collected dataset.
	
	\subsection{Preprocessing} \label{sec:Preprocessing}
	
	\textbf{Face Alignment}:
	In order to align all face images in the dataset, a reference shape is extracted from a selected subset of 2,000 face images in the passport style photos, i.e. frontal faces without expressions.
	All face images in the AGFW dataset are then warped to the texture domain corresponding to this reference shape. The warping step aims to remove the effects of shape variations during the texture modeling step. Finally, we obtain the dense correspondence between all faces in the training data.
	The DLIB tool \cite{dlib09} is employed to extract 68 landmarks for each face and the Procrustes Analysis is used to align these face images.

	\textbf{Expression Normalization}: The expressions in the images of each age group are further normalized using the Collection Flow technique \cite{kemelmacher2012collection}.
	
	\begin{figure*}[!t]
		\begin{center}
			\includegraphics[width=17.5cm]{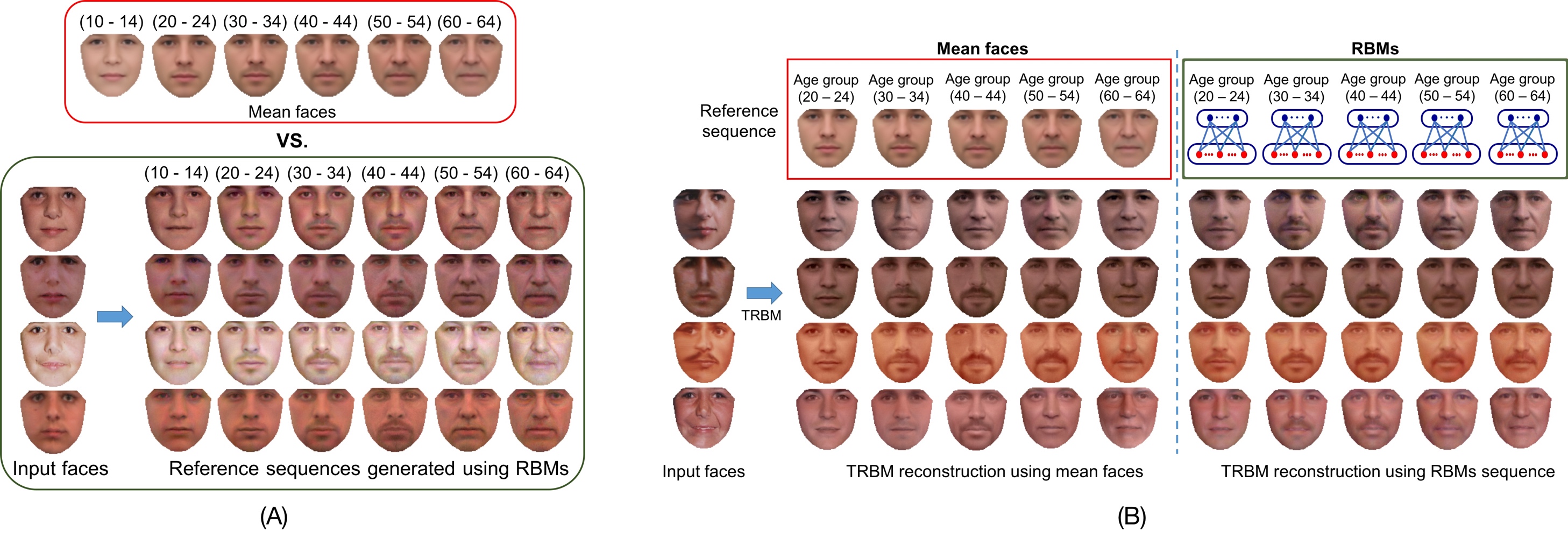}
		\end{center}
		\caption{A comparison between (A) two approaches to generate reference sequences and (B) synthesized aging faces using these two reference sequences. Faces in the red box: the sequence of mean faces in several age groups. Faces in the green box: reference faces generated by transferring features among RBMs of these age groups.
			Given input images in the age range of 10-14, our system automatically synthesizes a sequence of age-progressed images in various age ranges respectively.
			 }
		\label{fig:TRBM_syn_meanseq}
	\end{figure*}
	\subsection{Reference Sequence Generation} \label{RefSeqGeneration}
	This section presents how to generate the set of reference faces that are mainly different in age conditions.
	
	\textbf{Baseline}:
	A straightforward approach to construct the reference sequence is to order the mean faces of all age groups chronologically.
	The advantage of using mean faces is that several variations such as identity, occlusion can be removed. However, due to the averaging property, the aging variation is also smoothed out in the mean faces.
	Therefore, mean faces usually look younger than those from their own age groups.
	Moreover, it is noted that the lighting presented in the mean faces could be remarkably different from that of the input face. Figure \ref{fig:TRBM_syn_meanseq}(A) shows the unmatched tones between the sequence of mean faces and the input faces.

	\textbf{Our Improvement using RBMs}:
	
	Given an input face $I$ at a particular age, instead of using the set of mean faces in all age groups as the reference sequence, a set of RBMs is constructed to model faces in different age groups. The high-level features are then transferred among RBMs to generate the reference faces for $I$.
	
	In particular, for each age group $k$, all images collected at that age group are used to construct an RBM to model the distributions of texture features presented in this age group. Since the texture data is real-valued, the Gaussian-Bernoulli RBM (GRBM) is employed.
	Once RBMs of all age groups are constructed, given an input face image, its high-level features are first extracted using the RBM of the corresponding age group. These features are then transferred to the hidden layers of other RBMs to reconstruct the faces of other age groups. Gibbs sampling technique is used for this reconstruction stage.
	
	There are several advantages of using RBMs in this step.
	Firstly, RBMs can help to model faces in more details comparing to mean faces.
	Secondly, since each RBM is built for a particular age group, it has the ability to generalize the faces with specific aging features. Therefore, transferring the high-level features between RBMs can generate new faces that consist of both original subject and new aging features.
	Thirdly, the lighting has implicitly corrected during the reconstruction process.
	Figure \ref{fig:TRBM_syn_meanseq}(A) illustrates the sequence of mean faces and the RBMs reconstructions by transferring features in six age groups.
	
	\subsection{Modeling the Aging Transformation via TRBM}
	
	In order to learn the aging transformation between faces in the sequence, we employ a TRBM with	Gaussian visible units. As illustrated in Figure \ref{fig:TRBM_whole_system}(A), the model consists of two sets of visible units (i.e. $\mathbf{v}^t, \mathbf{v}^{t-1}$) encoding the texture of current face at age group $t$ and previous face at age group $t-1$; and a set of binary hidden units $\mathbf{h}^t$ that are latent variables. In addition, the faces in reference sequence, $\mathbf{s}^{<=t} = \{\mathbf{s}^t, \mathbf{s}^{t-1}\}$, at age group $t$ and $t-1$ are also incorporated by the connections to both hidden and visible units.
	
	The energy of the joint configuration $\{\mathbf{v}^t, \mathbf{h}^t\}$ is formulated as follows.
	\small
	\begin{equation} \label{eq:TRBMEnergy}
		\begin{split}
			E(\mathbf{v}^t,\mathbf{h}^t|\mathbf{v}^{t-1}, \mathbf{s}^{<=t};\boldsymbol{\theta})=&\sum_{i}{\frac{(v_i^t-b_i^t)^2}{2\sigma^2_i}} - \sum_j h_j^t a^t_j\\
			&-\sum_{i,j} {\frac{v_i^t} {\sigma_i} W_{ij} h_{j}^t}
		\end{split}
	\end{equation}
	\normalsize
	where $\boldsymbol{\theta} = \{\mathbf{W, A,B,P,Q}, \boldsymbol{\sigma}^2, \mathbf{b}^t, \mathbf{a}^t \}$  are the model parameters.
	In particular, $\{\mathbf{W, A,B,P,Q} \}$ are the weights of connections as illustrated in Figure \ref{fig:TRBM_whole_system}(A); $\{\boldsymbol{\sigma}^2, \mathbf{b}^t, \mathbf{a}^t \}$ are the variance, bias of the visible units and bias of the hidden units, respectively.
	Notice that the form of this energy function is very similar to the original form of an RBM. However, the bias terms are redefined as:
	\small
	\begin{eqnarray}
		b_i^t &=& b_i + B_i \mathbf{v}^{t-1} + \sum_l P_{li}\mathbf{s}_l^{<=t}\\
		a_j^t &=& a_j + A_j \mathbf{v}^{t-1} + \sum_l Q_{lj}\mathbf{s}_l^{<=t}
	\end{eqnarray}
	\normalsize
	where $l$ is the index of reference faces in sequence $\mathbf{s}^{<=t}$.
	
	The probability of $\mathbf{v}^t$ assigned by the model is given by
	\small
	\begin{equation}
		\begin{split}
			p(\mathbf{v}^t | \mathbf{v}^{t-1},\mathbf{s}^{<=t};\boldsymbol{\theta} ) &= \sum_{\mathbf{h}^t} p(\mathbf{v}^t, \mathbf{h}^t | \mathbf{v}^{t-1},\mathbf{s}^{<=t};\boldsymbol{\theta}) \\
			&= \frac{1}{Z} \sum_{\mathbf{h}^t}  e^{-E(\mathbf{v}^t,\mathbf{h}^t|\mathbf{v}^{t-1}, \mathbf{s}^{<=t};\boldsymbol{\theta})}
		\end{split}
	\end{equation}
	\normalsize
	where $Z$ is the partition function. The probability of a sequence with $T$ faces given the first face and the reference sequence $\mathbf{s}^{1:T}$ is defined as Eqn. (\ref{eqn:TRBM_seqProbs}).
	\small
	\begin{equation} \label{eqn:TRBM_seqProbs}
		p(v^{2:T} | \mathbf{v}^{1},\mathbf{s}^{1:T};\boldsymbol{\theta}) = \prod_{t=2}^T p(\mathbf{v}_t | \mathbf{v}^{t-1},\mathbf{s}^{<=t};\boldsymbol{\theta})
	\end{equation}
	\normalsize
	
	The conditional distributions over $\mathbf{v}^t$ and $\mathbf{h}^t$ are given as
	\small
	\begin{equation}
		\begin{split}
			p(h_j^t = 1|\mathbf{v}^t, \mathbf{v}^{t-1},\mathbf{s}^{<=t}) &= \sigma \left(\sum_i W_{ij}\frac{v_i^t}{\sigma_i} + a_j^t\right)\\
			v_i^t |\mathbf{h}^t, \mathbf{v}^{t-1},\mathbf{s}^{<=t} &\sim\mathcal{N} \left(\sigma_i\sum_j W_{ij}h_j^t + b_i^t, \sigma_i^2\right)
		\end{split}
	\end{equation}
	\normalsize
	
	\noindent
	\textbf{Model Properties}:
	With this structure, two types of information can be learned from the model:
	\begin{enumerate}[leftmargin=*,topsep=0.5pt]
		\itemsep0em
		\item The temporal information presented in the relationship between previous face $\mathbf{v}^{t-1}$ and the current face $\mathbf{v}^t$.
		\item The aging information provided by the reference sequence. This type of information acts as guidance information enforcing the model to learn the aging differences rather than other variations.
	\end{enumerate}
	Moreover, in order to transfer the information between faces, both linear and nonlinear interactions are employed in this model. In particular,  $\mathbf{v}^{t-1}$ and $\mathbf{v}^t$ are connected via two pathways: (1) the linear and direct connections using weight matrix $\mathbf{B}$; and (2) the nonlinear connections through the latent variables $\mathbf{h}^t$ with the weight matrices $\mathbf{A}$ and $\mathbf{W}$.
	Similar to the relationship between $\mathbf{v}^t$ and $\mathbf{s}^{<=t}$ , the direct (with weight matrix $\mathbf{P}$) and indirect (with weights $\mathbf{Q}$ and $\mathbf{W}$) connections allow both linear and nonlinear interactions. 
	Notice that except the undirected connections between hidden units $\mathbf{h}^t$ and visible units $\mathbf{v}^t$, all connections are directed.

	\noindent
	\textbf{Model Learning}:
	The learning process is to find the model parameters that maximize the log-likelihood:
	\small
	\begin{equation} \label{Eqn:ModelLearningMaximizeLikelihood}
		\theta^* = \arg \max_{\theta} \sum_{t=2}^{T} \log {p(\mathbf{v}_t | \mathbf{v}^{t-1},\mathbf{s}^{<=t};\boldsymbol{\theta}) }
	\end{equation}
	\normalsize
	The optimal parameter values can then be obtained via a gradient descent procedure given by
	\small
	\begin{equation}
		\frac{\partial}{\partial \theta}\mathbb{E}\left[\log p(\mathbf{v}_t | \mathbf{v}^{t-1},\mathbf{s}^{<=t};\boldsymbol{\theta})  \right]=\sum_{t=2}^T \mathbb{E}_{\text{data}}\left[\frac{\partial E}{\partial \theta}\right]-\mathbb{E}_{\text{model}}\left[\frac{\partial E}{\partial \theta}\right]
	\end{equation}
	\normalsize
	where $\mathbb{E}_{\text{data}}\left[ \cdot \right]$ and $\mathbb{E}_{\text{model}}\left[ \cdot \right]$ are the expectations with respect to data distribution and distribution estimated by the TRBM model.
	The Contrastive Divergence technique \cite{hinton2002training} is used for the learning process.

	\subsection{RBM based Wrinkle Modeling}
\begin{figure}[!t]
		\begin{center}
			\includegraphics[width=7.5cm]{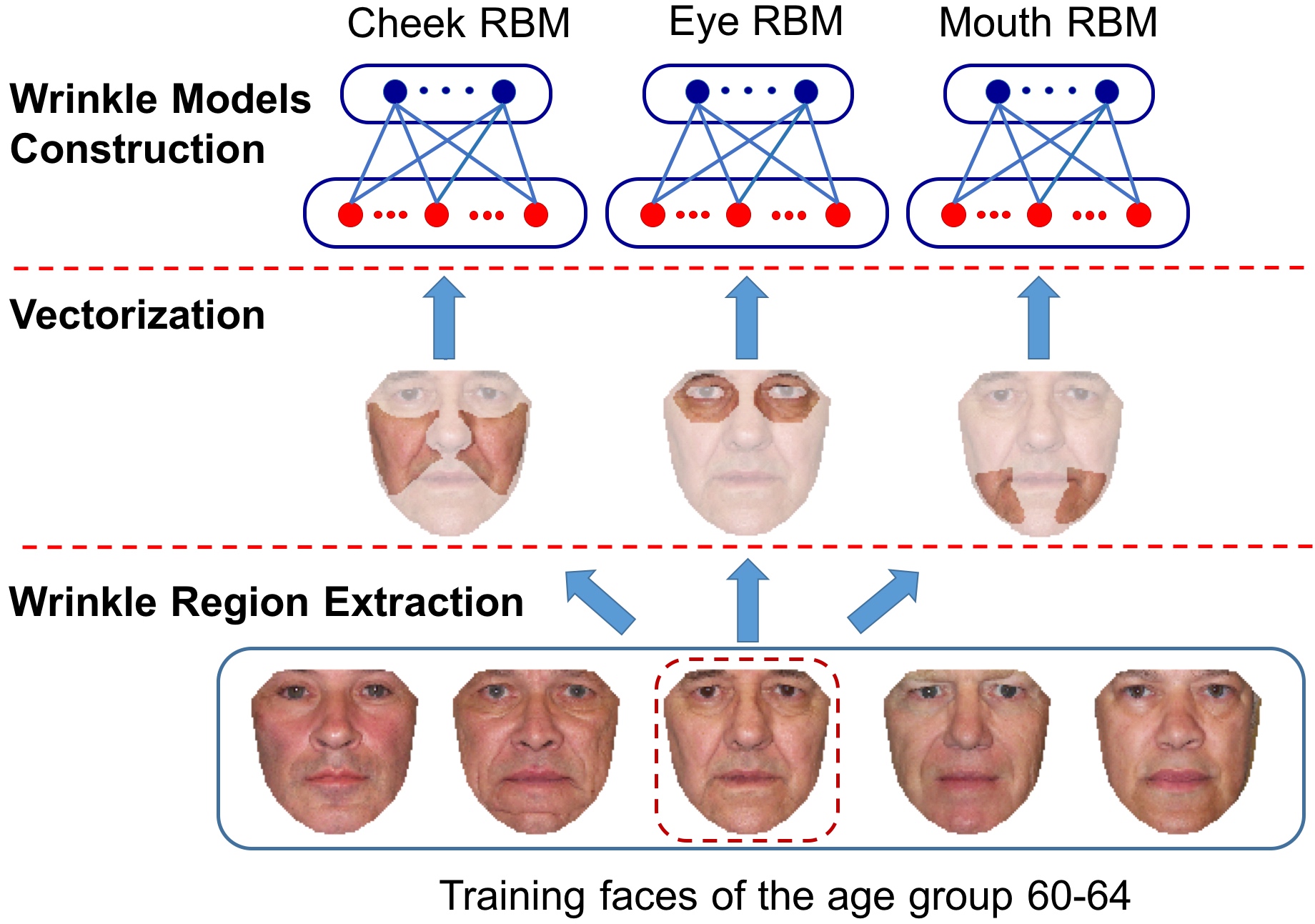}
		\end{center}
		\caption{Wrinkle Model Construction Steps.}
		\label{fig:WrinkleModelConstruction}
	\end{figure}

	Since facial muscles play an important role on the changes of wrinkle appearance during aging process, we  make use of the anatomical evidence for wrinkles enhancement.
	In particular, inspiring from the analysis on the behaviors of facial muscles \cite{ramanathan2008modeling},  we select the muscles that are more relevant to  wrinkle appearance and use their physical positions to extract the wrinkle subregions from the face image.
	Three chosen subregions are shown in Figure \ref{fig:WrinkleModelConstruction}.
	A set of RBMs is then employed to learn the distributions of wrinkle appearance for every age group.
	
\begin{figure}[t]
		\begin{center}
			\includegraphics[width=7.6cm]{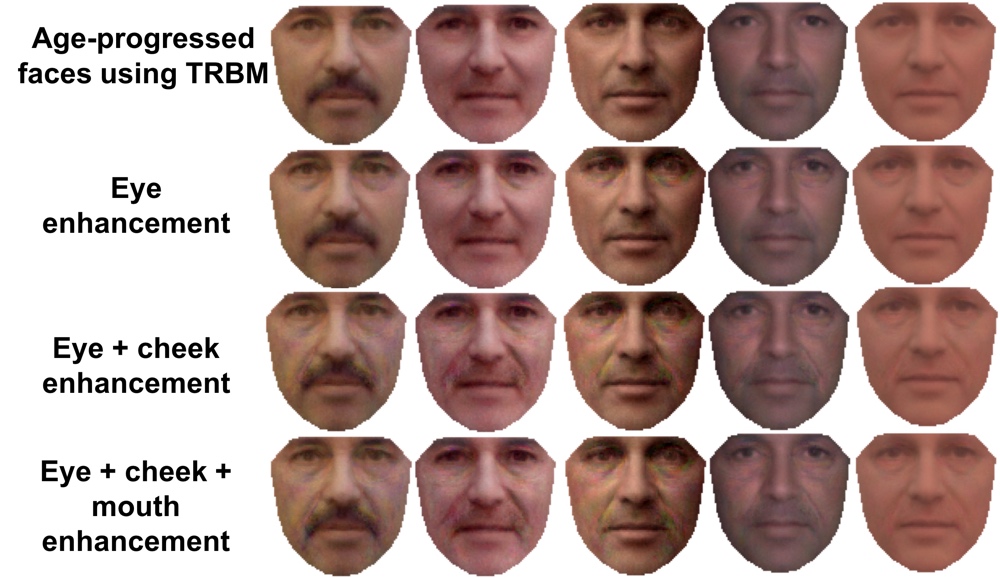}
		\end{center}
		\caption{Wrinkle Enhancement. From top to bottom: the synthesized images from the previous step, the results after enhancing eye; eye and cheek; eye, cheek and mouth regions.}
		\label{fig:WrinkleEnhancement}
	\end{figure}

	Once RBMs for all subregions and age groups are learned, the wrinkles are enhanced via a two-step process:
	(1) Generating the wrinkles through a Gibbs sampling process with the learned distributions; and
	(2) Wrinkle rendering by blending the generated wrinkles with the synthesized faces obtained from the TRBM based texture progression step.
	The Poisson blending technique \cite{perez2003poisson} is used for seamless fusion results.
	Figure \ref{fig:WrinkleEnhancement} shows the wrinkles enhancement results in three wrinkle regions.
	
	\subsection{Shape Adjustment}
	To further take into account the changes of shape during aging process, for each age group, we compute the average face shape using the same pipeline as in Section \ref{sec:Preprocessing} with the AGFW dataset.
	Then the synthesized faces obtained from the previous step are warped to the corresponding face shapes for the final age-progressed result.
	
	\section{Experimental Results}
	In this section, we evaluate the efficiency and flexibility of our proposed system in both age progression and regression applications.
	We next demonstrate the generality and robustness of our model with ``in the wild'' data.
	
	\subsection{Databases}
	For the training phase, we use two databases: the AGFW dataset collected as presented in section \ref{sec:DataCollection} and a subset of the Cross-Age Celebrity Dataset (CACD) \cite{chen14cross}. Then two public face aging databases: FG-NET \cite{fgNetData} and MORPH \cite{ricanek2006morph} are employed for evaluation.
	
	\textbf{Cross-Age Celebrity Dataset (CACD)} provides a large-scale dataset with 163446 images and the age ranging from 14 to 62. This dataset is collected from the Internet using keywords formed by the names of 2000 celebrities and the year (i.e. from 2004 to 2013).
	The annotations for this database are limited with 16 landmarks.
	
	\textbf{FG-NET} contains 1002 face images of 82 subjects with the age ranging from 0 to 69. In addition, each facial image is annotated with 68 landmark points.
	
	\textbf{MORPH} provides a large-scale dataset with two albums of passport style images. The MORPH-I includes 1690 images from 515 subjects and the age ranges from 15 to 68. The MORPH-II contains 55134 photos of 13000 subjects. In our experiments, MORPH-I is used for evaluation.
	
	\begin{figure}[!t]
		\begin{center}
			\includegraphics[width=7.4cm]{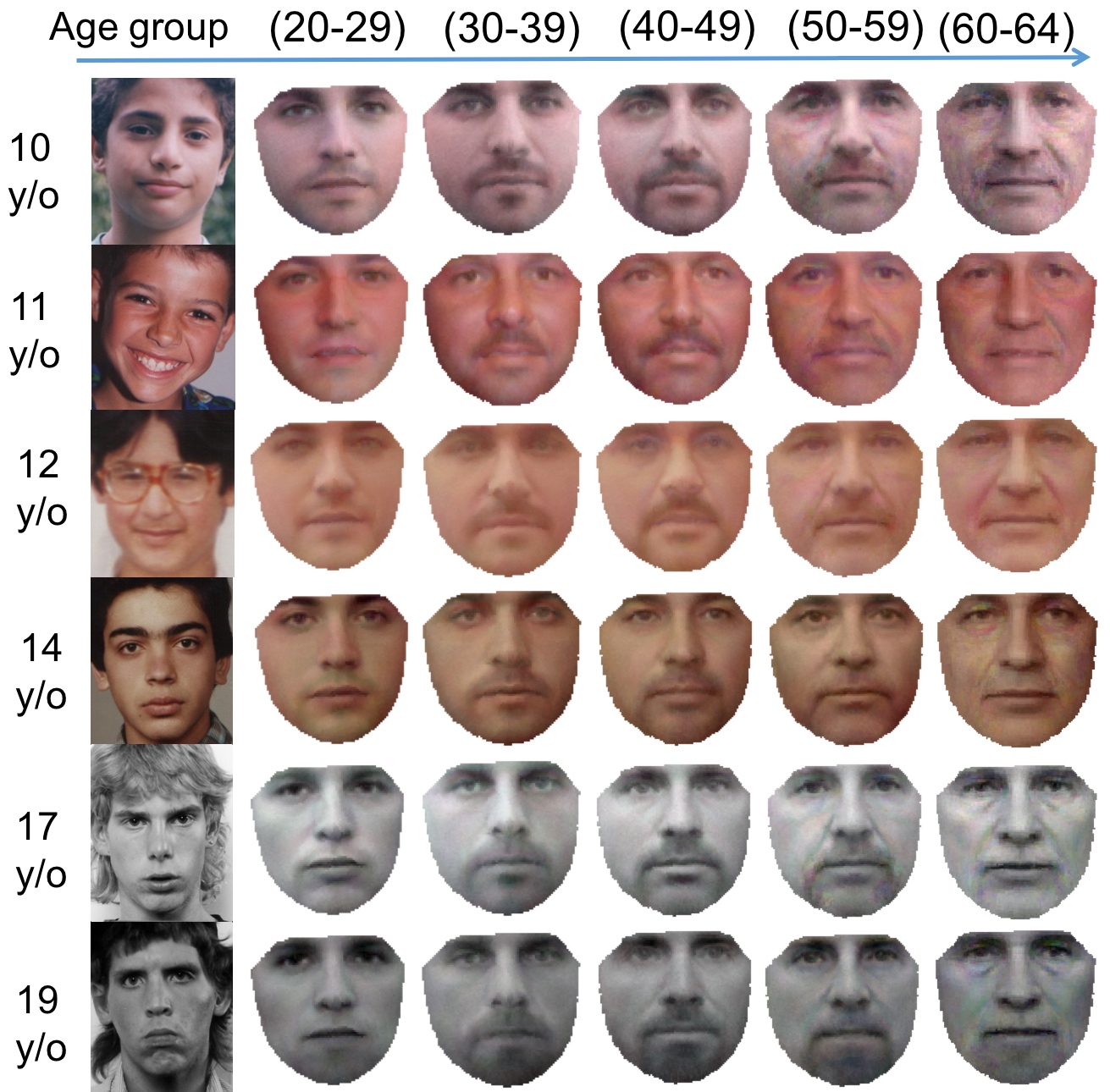}
		\end{center}
		\caption{Age progression results. Given an input image in age range 10-19, the system automatically reconstructs age-progressed images in various age ranges. 
		}
		\label{fig:TRBM_syn_RBMseq}
	\end{figure}
	\begin{figure*}[t]
		\begin{center}
			\includegraphics[width=12cm]{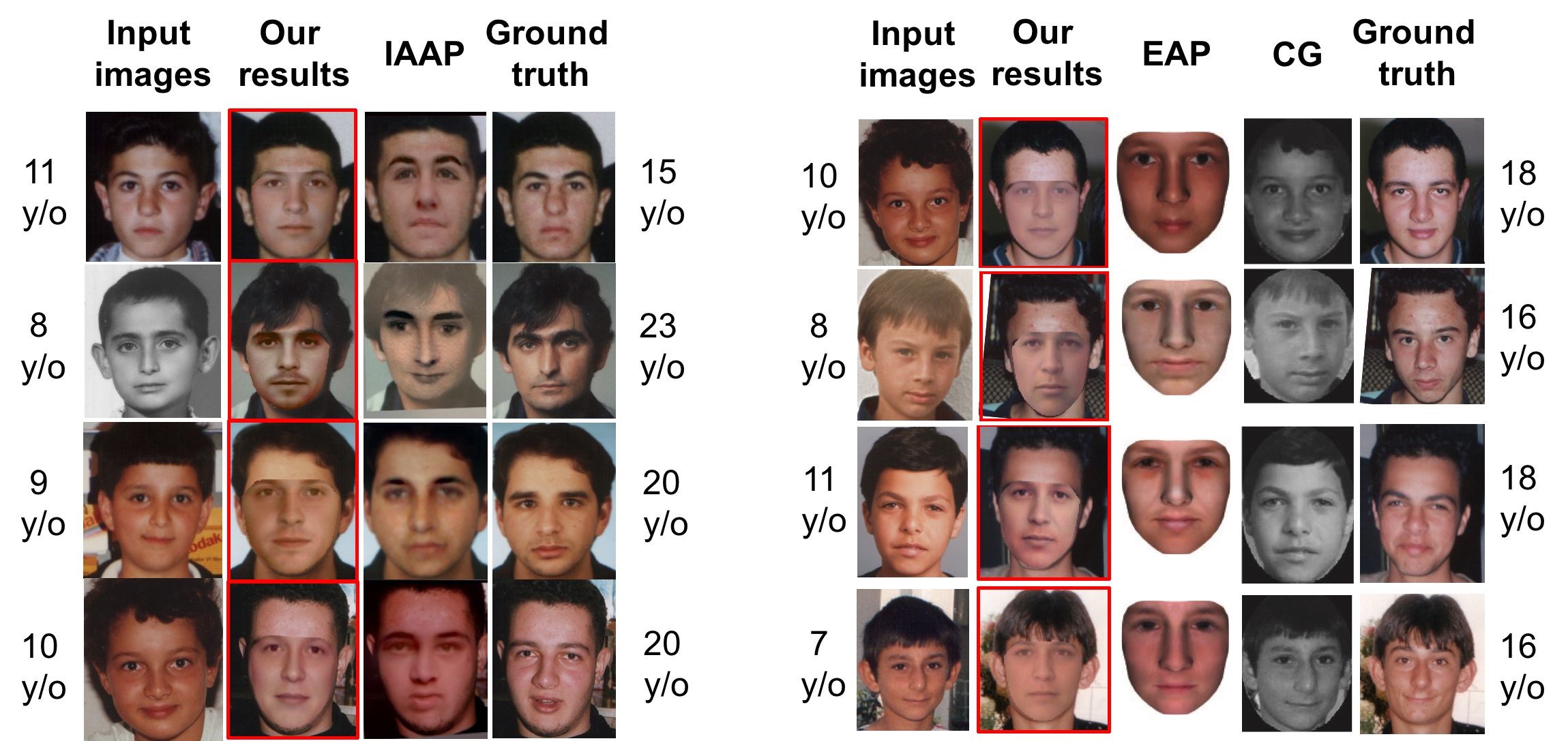}
		\end{center}
		\caption{Comparisons between our appproach and other age progression approaches: IAAP \cite{kemelmacher2014illumination}, EAP \cite{shen2011exemplar} and CG \cite{ramanathan2006modeling}.
		}
		\label{fig:AgeProgressionComparisons}
	\end{figure*}
	\begin{figure}[t]
		\begin{center}
			\includegraphics[width=7.5cm]{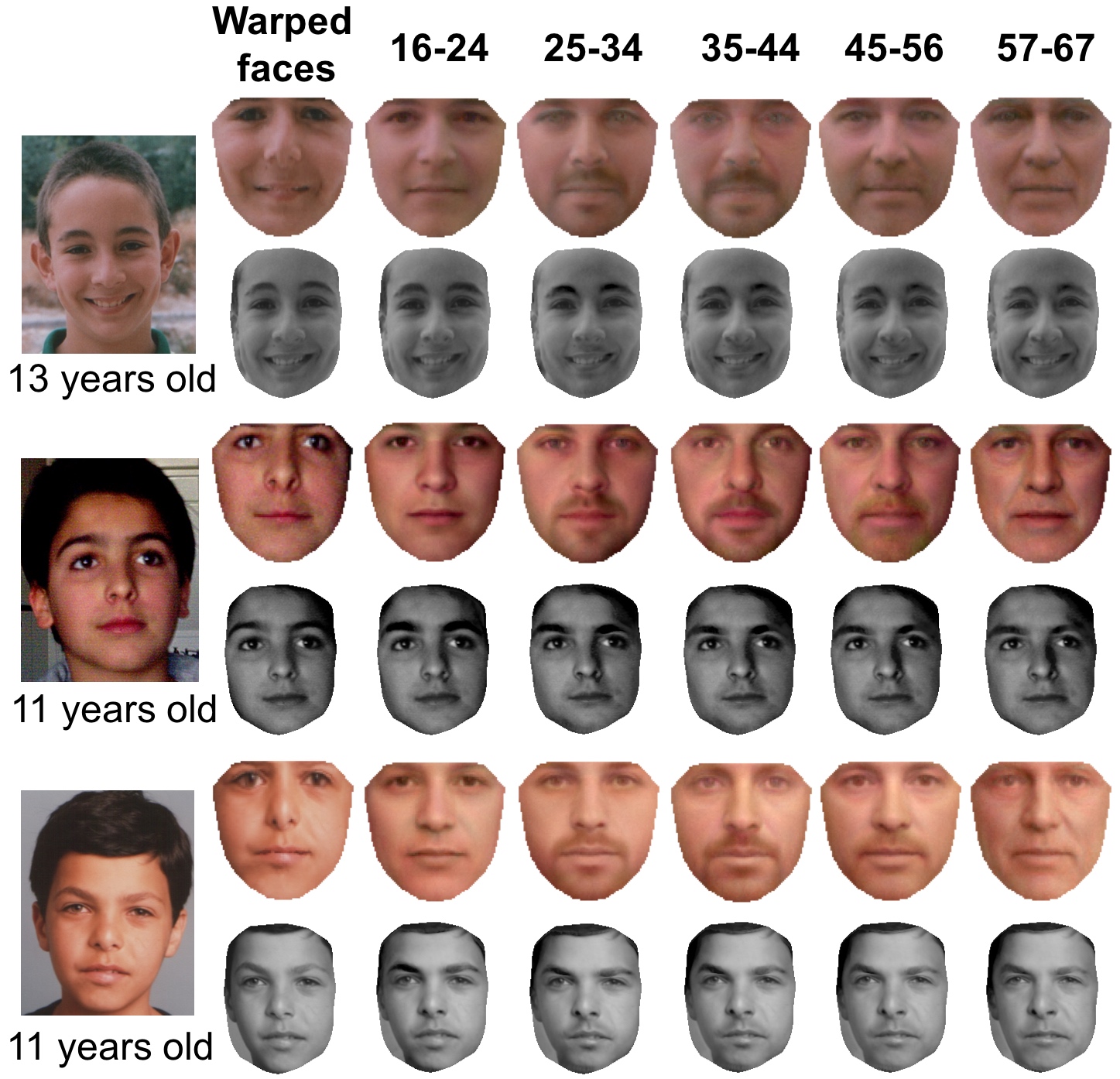}
		\end{center}
		\caption{Comparisons between our approach and IAAP \cite{kemelmacher2014illumination}. For each case, the input face image (1st column) is aligned and normalized to frontal face (2nd column). From the 3rd to the 7th column: the progressed images corresponding to several age groups using our approach (the row above) and IAAP (the row below).
		}
		\label{fig:AgeProgressionComparisonsSeq}
	\end{figure}
	
	\subsection{Age Progression}
	In order to train the RBMs for reference sequence generation, the AGFW dataset is decomposed into 11 age groups with the age span of 5  (i.e. age 10-14, 15-19, ..., 60-64). On average, each age group consists of 1700 images. These images are then used for constructing the set of RBMs as represented in Section \ref{RefSeqGeneration}.
	For training the TRBM based age progression component, we select a subset of 572 celebrities from  the CACD dataset and also classify their images into 11 age groups with the age span of 5.
	Then for each person, one image per age group is randomly selected. This process results in a training data with 572 sequences.
	Since the images are collected from 2004 to 2013, the longest sequence consists of only three images.
	
	All training images are then aligned and normalized as presented in section \ref{sec:Preprocessing}.
	The size of the normalized image is set to $95 \times 95$ pixels based on the reference shape generated in the alignment step.
	The TRBM based age progression model is then employed to learn the aging transformation between faces.
	After all components are trained, we run our system on every face over 10 years old of FG-NET and MORPH databases.
	Figure \ref{fig:TRBM_syn_RBMseq} illustrates the age-progressed faces reconstructed by our model.
	Notice that both FG-NET and MORPH databases are not part of our training data.
	
	Our age-progressed sequences are also compared with the recent age progression work, Illumination-Aware Age Progression (IAAP) \cite{kemelmacher2014illumination} against FG-NET database in Figure \ref{fig:AgeProgressionComparisonsSeq}. From these sequences, one can see that IAAP approach synthesizes very similar faces among different age groups.
	Moreover, since the texture difference between average faces is used as the main source for aging process, the synthesized faces usually look younger than those from their own age groups.
	Meanwhile, more nonlinear aging features in each age group are still kept in the reconstructed results of our approach.
	In addition, one can easily see that our age-progressed sequences are able to better reflect the face changes during the aging process (i.e. the appearance of beard in the middle stages and wrinkle in the later stages).
	For further evaluations, we compare our proposed model with other approaches including IAAP; Exemplar based Age Progression (EAP) \cite{shen2011exemplar} and Craniofacial Growth (CG) model \cite{ramanathan2006modeling} in Figure \ref{fig:AgeProgressionComparisons}. The ground truth images are also provided for comparisons.
	It should be noted that since our model is trained using the collected data with ages ranging from 10 to 64, in cases where the IAAP uses input images at ages less than 5, we choose images of the same individuals with age close to 10 as input for our system.
	
	\begin{figure}[!t]
		\begin{center}
			\includegraphics[width=7.9cm]{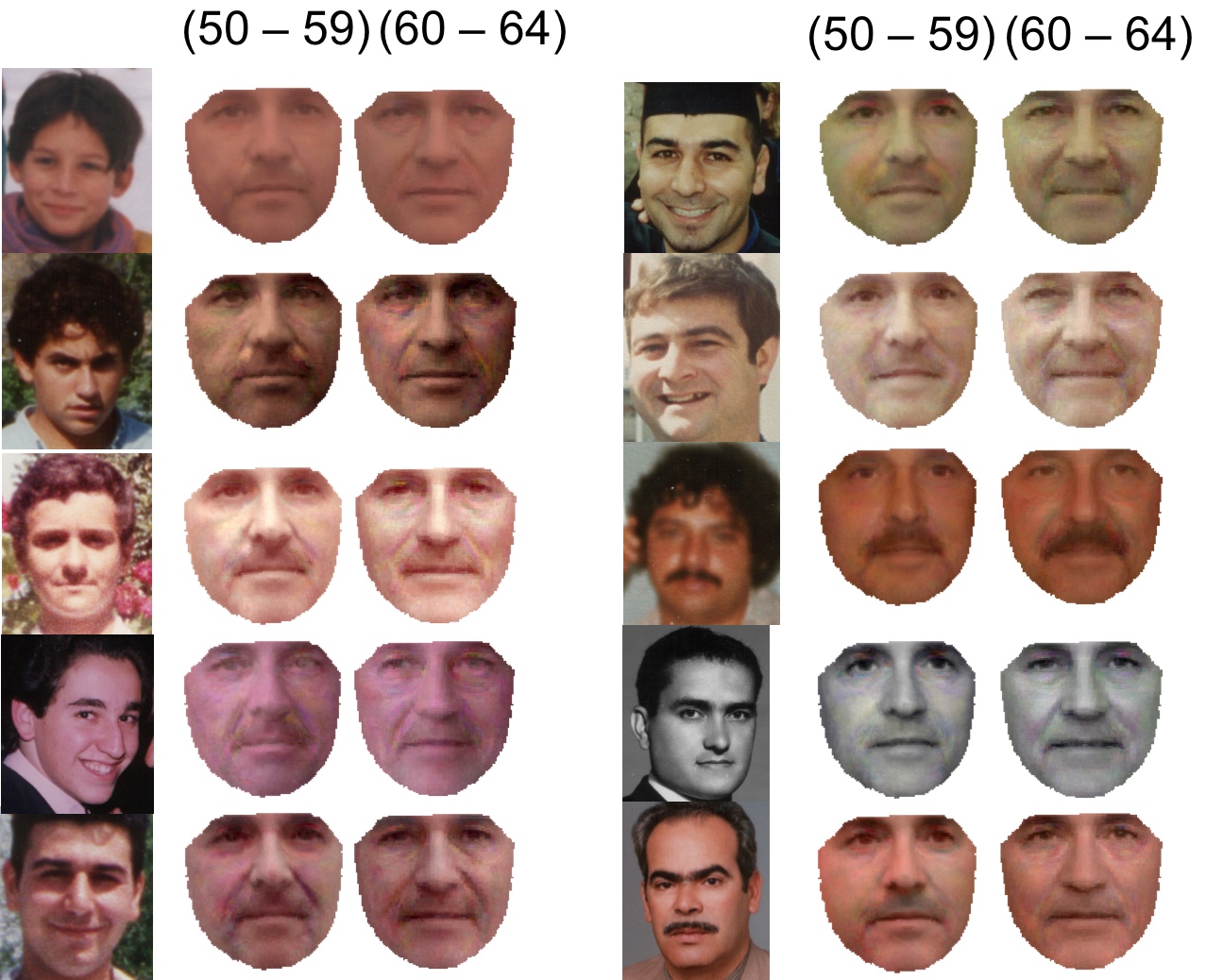}
		\end{center}
		\caption{Age progression ``in the wild'' with other variations in the input images such as poses, illuminations, expressions.
		}
		\label{fig:AgeProgressionInTheWild}
	\end{figure}
	
\subsection{Age Progression ``in the Wild''}
	In order to validate the robustness of our model, in this experiment, we focus on input images that include different variations such as poses, expressions, illuminations. Blurry images are also considered.
	Figure \ref{fig:AgeProgressionInTheWild} illustrates age-progressed images that are automatically reconstructed by our model.
	From these results, one can see that although other non-linear variations also present in the input images, remarkable results can still be achieved by our model in terms of fine aging details without any quality reduction.
	
\subsection{Age Regression}
\begin{figure}[t]
		\begin{center}
			\includegraphics[width=8cm]{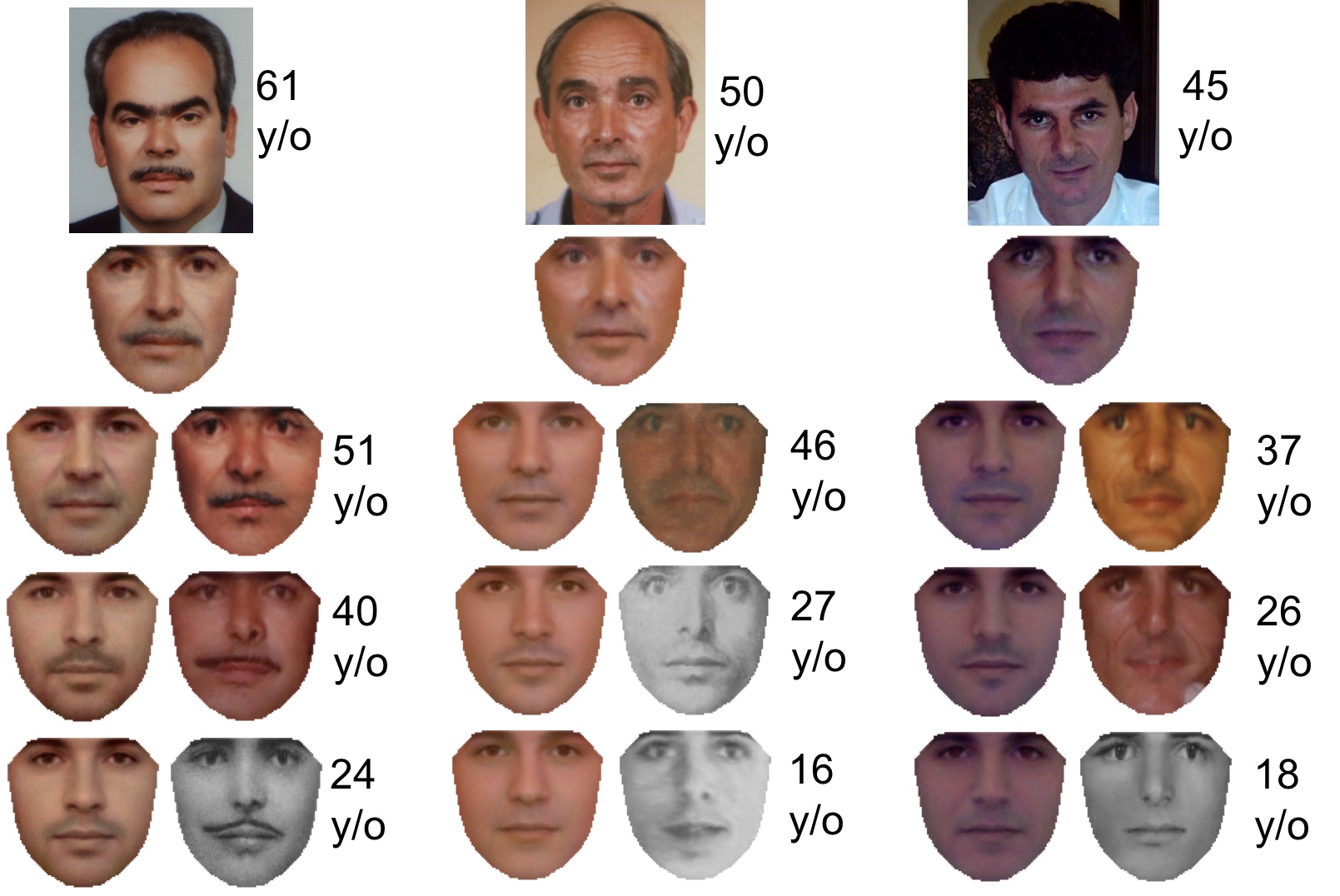}
		\end{center}
		\caption{Age regression results. For each case, the input image (1st row) is normalized to frontal face (2nd row). From the 3rd row to 5th row: the age-regressed images generated by our model (left) and the ground truth images with the corresponding ages (right).
		}
		\label{fig:AgeRegression}
	\end{figure}

	We next emphasize the flexibility of our proposed model by evaluating its capability to generate the younger faces of an individual given his/her current appearance.
	The results of this application can be easily obtained using our model by simply keeping the same training process as in previous experiments except the training sequences are reversed. The faces at younger ages are represented in Figure \ref{fig:AgeRegression}.

\subsection{Automatic Age Estimation} \label{sec:AgeEstimation}
	One challenge of the face data ``in the wild'' comes from the age labels of the input images. In most cases, this information is incorrect or unavailable. Thus, it causes lots of difficulties for age progression process in later stage.
	Far apart from previous age progression systems,  the effectiveness and scalability of our proposed model is further increased by integrating an age estimation system to the proposed framework.
	In this way, given a face image, our system can do age progression without any further information.
	
	Besides some other previous age estimation approaches \cite{duong2015beyond, juefei2011investigating, luu2011kernel, luu2011contourlet}, in this work, we re-implement the method in \cite{luu2009age} which is among the state-of-the-art age estimators reported in \cite{NISTReport}.
	Moreover, this approach is modified with three-group classification in the first step (youths, adults, and elders) before constructing three Support Vector Regression (SVR) based aging functions.
	In order to train this age estimator, we randomly select 802 images from FG-NET and 1000 images from MORPH as the training data.
	The remaining images of these two databases are used for testing.
	The Mean Absolute Errors (MAEs) achieved are 5.86 years for FG-NET and 4.84 years for MORPH.
	By incorporating this age estimator to our age-progression framework, the need for age label is alleviated and, therefore, making the whole framework fully automatic.

	\begin{table}[t]
		\caption{The MAEs (years) of Age Estimation System on Ground Truth and Age-progressed Results}
		\label{table_MAE_age_estimation_synthesized}
		\centering
		\begin{tabular}{|c|c|c|}
			\hline
			Inputs & Dataset & MAEs \\
			\hline
     		\hline
			Ground Truth faces (set A)& FGNET&  5.89\\
			\hline
			Synthesized faces (set B)& FGNET& 5.96 \\
			\hline
			IAAP 's synthesized faces (set B')& FGNET& 6.29 \\
			\hline
			\hline
			Ground Truth faces (set C)&MORPH& 4.84 \\
			\hline
			Synthesized faces (set D)&MORPH&5.17 \\
			\hline
		\end{tabular}
	\end{table}
	\subsection{Age Accuracy of Age-progressed Results}
	This section illustrates the accuracy of our synthesized results in term of age perceived. In other words, this experiment aims at assessing whether the age-progressed faces are perceived  to be at the target ages.
	In this evaluation, the trained age estimation system in the previous experiment is adopted to compare the accuracies on the ground-truth and age-progressed faces.
	From the testing set of FG-NET database, we select all images above 10 years old and consider them as the ground truth images. This forms the set A consisting of 135 images.
	Each photo of an individual in set A is then progressed to the later ages where the ground truth faces are available. This process results in the set B of 194 age-progressed images.
	In order to compare with IAAP method, we apply this process using IAAP and obtain the set B'.
	For a large scale evaluation, we further generate a test set using MORPH database.
	Let the test set of MORPH as in section \ref{sec:AgeEstimation} be set C. For each individual in the testing data, we synthesize four aged images accross three decades. This gives us 1421 images that compose set D.
	The MAEs of the age estimation system on these test sets are listed in Table \ref{table_MAE_age_estimation_synthesized}. These results show that the age estimation accuracies of our age-progressed images are comparable to those of ground truth images. Therefore, our proposed model is able to generate the age-progressed faces at the target ages.

	\section{Conclusion}
	This paper has developed a novel deep model based approach for face age progression that  can operate ``in the wild''.
	With the deep structured models for both face representation and aging transformation modeling, the proposed model can efficiently capture the non-linear aging changes as well as robustly handle other variations such as pose, expressions, and illuminations.
	The aging rules in terms of wrinkle appearance and geometric constraints are also taken into account for more consistent progression results.
	Experimental results in both age progression and age regression applications have shown the efficiency, generality and flexibility of our proposed model.

{\small
\bibliographystyle{ieee}
\bibliography{egbib}
}

\end{document}